\documentclass{article}

\usepackage[preprint]{neurips_2026}

\usepackage[utf8]{inputenc} 
\usepackage[T1]{fontenc}    
\usepackage{hyperref}       
\usepackage{url}            
\usepackage{booktabs}       
\usepackage{amsfonts}       
\usepackage{nicefrac}       
\usepackage{microtype}      
\usepackage{xcolor}         
\usepackage{colortbl}
\usepackage{subcaption}
\usepackage{tcolorbox}
\usepackage{enumitem}
\usepackage{siunitx}
\usepackage{multirow}
\usepackage{amsmath}
\usepackage{threeparttable}
\usepackage{cleveref}
\crefname{section}{Section}{Sections}
\Crefname{section}{Section}{Sections}
\crefname{table}{Table}{Tables}
\Crefname{table}{Table}{Tables}
\crefname{figure}{Figure}{Figures}
\Crefname{figure}{Figure}{Figures}
\usepackage{url}
\usepackage{booktabs}
\usepackage{graphicx}
\usepackage{wrapfig}
\usepackage{xspace}
\usepackage{xcolor}
\usepackage[cjk]{kotex}
\usepackage[font=small]{subcaption}
\usepackage[font=footnotesize]{subcaption}
\usepackage{graphicx} 
\usepackage{lipsum}
\usepackage{tcolorbox}
\newtcolorbox{messagebox}{
  colback=gray!5,       
  colframe=gray!50,     
  boxrule=0.5pt,        
  arc=2pt,              
  left=4pt, right=4pt, top=4pt, bottom=4pt, 
  coltitle=black,
  fonttitle=\bfseries,
  before=\vspace{0.5em}, 
  after=\vspace{0.5em}   
}
\usepackage[hang,flushmargin]{footmisc}

\newcommand{\ie}{\textit{i.e.,}\xspace}
\newcommand{\eg}{\textit{e.g.,}\xspace}

\usepackage[dvipsnames]{xcolor}

\newcommand\rethinks{\textit{rethink ``wait''s}\xspace}
\newcommand\recalls{\textit{recall ``wait''}s\xspace}
\newcommand\rethink{\textit{rethink ``wait''}\xspace}
\newcommand\recall{\textit{recall ``wait''}\xspace}
\newcommand\Rethink{\textit{Rethink ``wait''}\xspace}

\usepackage{soul}
\newcounter{takeawayonly}
\tcbuselibrary{skins}
\tcbuselibrary{breakable} 

\usepackage{circledtext}

\usepackage{tcolorbox} 

\newtcolorbox{chattemplatebox}[1][]{
  enhanced,
  breakable,
  colback=gray!3,
  colframe=black!15,
  boxrule=0.5pt,
  arc=2mm,
  left=1em, right=1em, top=0.8em, bottom=0.8em,
  fonttitle=\bfseries,
  title=Chat Template,
  #1
}

\title{\textit{Oops, Wait}: Discourse Tokens Matter \\in Reasoning Model}

\author{
 Jaehui Hwang \quad
 Byeongho Heo  \quad
 Sangdoo Yun \quad
 Dongyoon Han\\
 NAVER AI Lab
\\
\texttt{jaehui.hwang@navercorp.com}
}

\begin{document}

\maketitle

\begin{abstract}
Recent studies suggest that even data-efficient training with ($\simeq$1K) reasoning trajectories can induce non-trivial reasoning capabilities in large language models through post-training. Such training corpora often contain iconic tokens such as ``wait'', ``so'', and ``alternatively'', which frequently appear in reasoning trajectories and may play a role in this process. 
This paper focuses on characterizing observable token-level patterns in post-training and 
a case study of how data-efficient supervised fine-tuning (SFT) differs from, and falls short of, large-scale post-training. To this end, we first identify tokens that correlate with correct answers along reasoning trajectories across models and training setups. We then focus on the distribution and (functional) roles of the ``wait'' token to primarily study the model trained in a data-efficient manner compared with the counterpart. Our study finds that discourse tokens are associated with correctness and a reasoning accuracy jump, even in data-efficient SFT. This suggests data-efficient SFT can partially reproduce discourse-token patterns to mimic meaningful reasoning behavior, but the patterns are less aligned with high-confidence answer transitions than those from large-scale post-training.
\end{abstract}

\section{Introduction}
Large language models (LLMs) have recently made remarkable progress across a wide range of reasoning-intensive tasks \citep{gpt4, deepseek, qwq, qwen2.5, qwen3, mixtral, kimi}. 
Modern reasoning models explicitly generate a \textit{reasoning trajectory}, which contains a step-by-step thinking process \citep{cot, zeroshot-cot, gpt4, survey_longcot}. 
These trajectories improve human interpretability \citep{biologyllm}, help to refine reasoning ability during post-training \citep{deepseek, openthought, s1, limo}. Most importantly, training on reasoning trajectories leads to substantial gains on complex benchmarks.



Beyond reasoning performance, recent studies highlight the frequent emergence of discourse tokens\footnote{We define tokens such as ``wait'', ``so'', ``therefore'', and ``alternatively'' as \textit{discourse tokens} marking pivotal steps in problem solving and anchor reasoning structure and transitions.} within reasoning trajectories \citep{yang2025understanding, aha-moment,qian2025demystifying}. 
Likewise, in human language, discourse markers play a role in structuring arguments and using language fluently \citep{sun2013importance,castro2009use,huneety2023use,stab-gurevych-2017-parsing}. This suggests that how LLMs employ particular tokens during reasoning may be closely tied to reasoning ability. 
A few studies \citep{s1, alphaone, nowait, zhao2025activation} have investigated the use of discourse tokens to improve reasoning capabilities. 
In contrast, how these tokens are distributed and how their patterns relate to reasoning success across models and training setups remains underexplored.


In human language learning, \citet{sun2013importance} suggests that acquiring discourse markers is an essential step towards developing persuasive and logical argumentation. 
By analogy, models post-trained on reasoning trajectories may acquire and internalize token-level signals, especially the discourse tokens, related to reasoning ability. However, the reasoning training setting differs substantially from human learning. Humans develop such capabilities through repeated exposure across vast and diverse contexts, and LLMs are thus generally assumed to require millions of post-training examples to reliably induce comparable reasoning. Nevertheless, recent studies~\citep{limo,s1} have shown that post-training with only a few curated reasoning examples can enhance LLMs' reasoning ability. Particularly, \citet{s1} demonstrated even $\simeq$1K examples may suffice to elicit non-trivial reasoning behavior in LLMs - an unexpectedly strong effect given the limited supervision. A hypothesis of this success points to the discourse tokens 
that pervade training reasoning trajectories and may act as structural cues. However, while \citet{s1} suggests that data-efficient supervised fine-tuning (SFT) can induce reasoning behaviors, it remains underexplored how such behaviors relate to those observed in large-scale post-trained models. In particular, it is unknown whether these models exhibit similar discourse-level dynamics during reasoning.

In this paper, we focus on empirical analyses of discourse tokens in reasoning models and on a case study comparing data-efficient SFT with large-scale post-training.
We begin by identifying which tokens are most salient, how they are distributed, and how they correlate with reasoning success. We then examine what is distilled by small, well-curated reasoning traces, through data-efficient SFT. In particular, we compare the model trained on only $\sim$1K examples (\eg s1.1-32B~\citep{s1}) with the models trained on orders of magnitude more data. These include DeepSeek-R1-Distill-Qwen-32B~\citep{deepseek} with $\sim$800K examples and QwQ-32B~\citep{qwq}, which was trained on substantially larger but undisclosed data. We go beyond their reasoning performance to study what distinguishes these regimes and characteristic reasoning behaviors.  


Our study finds that (1) the discourse tokens are associated with correctness and a reasoning accuracy jump, (2) these correlations vary by training regime but are consistent across model scales, (3) data-efficient SFT could transfer signals using the discourse tokens like ``wait'' that connect to reasoning success under limited data; however, the discourse tokens' patterns are less aligned with high-confidence answer transitions, and therefore, its overall effectiveness remains questionable. 
Our analyses motivate exploratory experiments to validate our insights, including an ensemble method based on our token-level signals that improves \texttt{pass@1} from 10\%p to 13.4\%p across the models: \{R1-32B, QwQ-32B, s1.1-32B\}. Finally, we aim to focus on observable patterns and correlations between discourse tokens and reasoning capability.

\section{Related Work}

\noindent\textbf{Anthropomorphic expressions and discourse markers in reasoning.}
Recent studies have investigated how such markers appear in LLM reasoning. \citet{yang2025understanding} analyzed external signals interpreted as anthropomorphic expressions associated with ``aha-moments'', while \citet{aha-moment} discussed the emergence of reflective patterns such as ``wait'' or ``hmm'' as discourse markers in R1-style reasoning models. \citet{qian2025demystifying} further identified that thinking tokens correspond to peaks of mutual information within reasoning trajectories. 
These works suggest a link between linguistic discourse markers and reasoning behavior, but focus on individual models. In contrast, we examine how discourse tokens relate to both the success of data-efficient SFT models and their side effects.


\begin{figure*}[b]
\centering
\includegraphics[width=0.85\textwidth]{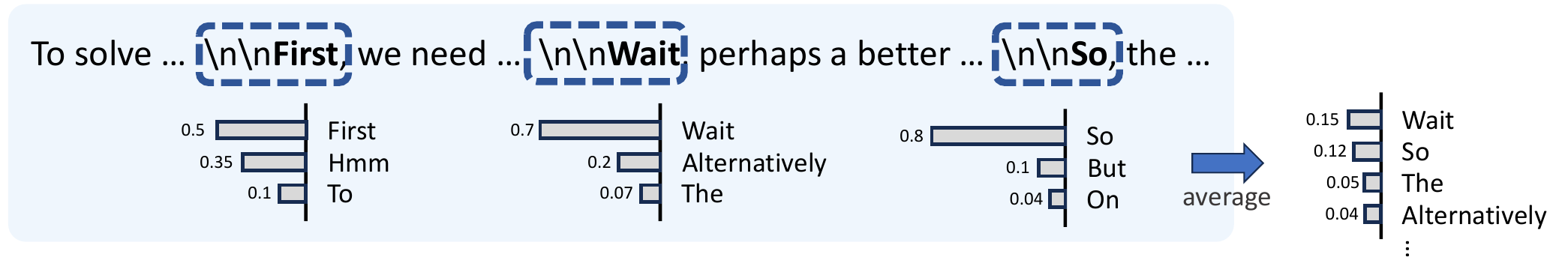}
\caption{\textbf{Overview of how token probabilities are collected.}
We extract next-token probability distributions specifically after ``\textbackslash{n}\textbackslash{n}'', which serve as natural segmentation points in the trajectories. We hypothesize that the first token plays a significant role in autoregressive generation.
}
\label{fig:trajectory_comparison}
\end{figure*}

\noindent\textbf{``Wait'' in reasoning of LLMs.}
\citet{s1} introduced test-time scaling via controlled reasoning extension with ``wait''. Building on this idea, \citet{zhao2025activation} linked ``wait'' to activation patterns and proposed an activation control method, while \citet{alphaone} modeled reasoning as a balance between slow and fast thinking mediated by the ``wait'' token. In parallel, \citet{nowait} reported that removing ``wait''-like tokens can even improve efficiency in some settings. While these studies highlight the importance of ``wait'' in shaping reasoning behavior, they primarily use it as control rather than as an analytical lens. Our work shifts the focus on how ``wait'' functions within reasoning trajectories for data-efficient SFT.


\section{Discourse Tokens in Reasoning}
\label{sec:3}
This section studies token-level reasoning trajectories. We quantify the contribution of candidate discourse tokens to reasoning success, identify the most effective tokens, and examine these effects across model scales and training strategies.

\subsection{Identifying discourse tokens}
We quantify the discourse token signals by computing token probabilities, defined as the average probability of tokens generated immediately after ``\textbackslash n\textbackslash n'', rather than relying on simple frequency counts. This allows us 1) to compare how tokens behave across correct and incorrect reasoning, across various models; 2) to examine their relation to model confidence. By examining these dimensions, we identify which tokens consistently act as markers of successful reasoning and which are associated with errors, providing a quantitative basis for understanding how training strategies and model scales shape reasoning ability.

\begin{table*}[t]
\centering
\footnotesize
\centering
\begin{minipage}[t]{0.48\linewidth}
\centering
\begin{tabular}{c r r r}
\toprule
\textbf{Token} & $\bar p_{true}(t)$ & $\bar p_{false}(t)$ 
& \multicolumn{1}{c}{$\Delta(t)$} \\
\midrule
I              & 4.02\%  & 1.30\%  & +2.72\%  \\
Therefore      & 4.14\%  & 1.60\%  & +2.54\%  \\
The            & 5.02\%  & 3.14\%  & +1.87\%  \\
Let            & 3.14\%  & 1.40\%  & +1.74\%  \\
First          & 2.50\%  & 0.82\%  & +1.68\%  \\
\midrule
If             & 0.61\%  & 0.96\%  & -0.34\%  \\
In             & 0.73\%  & 1.66\%  & -0.94\%  \\
But            & 3.74\%  & 7.23\%  & -3.49\%  \\
Alternatively  & 3.74\%  & 8.98\%  & -5.23\%  \\
Wait           & 15.40\% & 25.75\% & -10.34\% \\
\bottomrule
\end{tabular}
\subcaption{R1-32B}
\end{minipage}
\hfill
\begin{minipage}[t]{0.48\linewidth}
\centering
\footnotesize
\begin{tabular}{c r r r}
\toprule
\textbf{Token} & $\bar p_{true}(t)$ & $\bar p_{false}(t)$ 
& \multicolumn{1}{c}{$\Delta(t)$} \\
\midrule
Therefore      & 7.58\%  & 3.59\%  & +3.98\%  \\
So             & 5.57\%  & 3.74\%  & +1.84\%  \\
First          & 3.00\%  & 1.47\%  & +1.53\%  \\
I              & 1.74\%  & 0.70\%  & +1.04\%  \\
Now            & 2.03\%  & 1.19\%  & +0.84\%  \\
\midrule
This           & 0.39\%  & 0.76\%  & -0.37\%  \\
If             & 0.63\%  & 1.17\%  & -0.54\%  \\
In             & 0.49\%  & 1.40\%  & -0.91\%  \\
The            & 5.49\%  & 8.54\%  & -3.05\%  \\
Alternatively  & 7.81\%  & 17.75\% & -9.94\%  \\
\bottomrule
\end{tabular}
\subcaption{QwQ-32B}
\end{minipage}
\caption{\textbf{Discourse tokens with statistically significant differences}
(\textit{t}-test, $p<0.05$) 
between correct and incorrect trajectories for \textbf{DeepSeek-R1-distill-Qwen-32B (R1-32B)} and \textbf{QwQ-32B}. 
Tokens are sorted by $\Delta(t)$.}
\label{tab:deepseek_tokens}
\end{table*}

\noindent\textbf{Mean token probability over the reasoning trajectory.}
Let $X = \{x_1, x_2, \dots, x_T\}$ denote the token set in a trajectory, and let $I_X = \{i:x_i= \text{``\textbackslash n\textbackslash n''},\; x_i \in X \}$ be a set of positions where ``\textbackslash n\textbackslash n'' occurs.  
The average token probability for trajectory $X$ is
$p_X = \frac{1}{|I_X|}\sum_{i \in I_X} p(x_{i+1}),$
where $p(x_{i+1})$ denotes the probability assigned by the model to token $x_{i+1}$. We further compute aggregated statistics across multiple trajectories.  
Among all trajectories in a dataset $D = \{X_1, X_2, ... X_N\}$, we use two subsets: correct-answer $D_{true} = \{X_i: \mathrm{Answer}(X_i) = true\}$ and incorrect-answer $D_{false} = \{X_i: \mathrm{Answer}(X_i) = false\}$.
Specifically, we define the mean token probability over correct- and incorrect-answer trajectories as
\begin{equation}
\bar{p}_{*} = \frac{1}{|D_{*}|} \sum_{X \in D_{*}} p_X,
\label{eq:mtp}
\end{equation}
where ${*} \in \{true, false\}$ for correct $p_{true}$ or incorrect $p_{false}$ mean probabilities, respectively. \cref{fig:trajectory_comparison} illustrates the overall process of how the token probability is computed by collecting the softmax probabilities:  
we use the token following ``\textbackslash n\textbackslash n''  as a high-precision discourse boundary to reduce contamination from problem-specific tokens. 
Note that we verify in Appendix \ref{sec:app-2} that alternatives yield similar trends, albeit with weaker signal strength.



We detail the benchmarks and models in Appendix \ref{sec:app-1} and \ref{sec:app-2}. For each tested model, \textit{t}-test is performed on token probability distributions to assess whether they differ significantly between correct and incorrect reasoning.
In addition, we prompt the models to generate confidence estimates, enabling us to examine the interactions among correctness, confidence, and token-level probabilities.

\noindent\textbf{Identified discourse tokens and their relation to answer correctness.} Tables~\ref{tab:deepseek_tokens} present representative examples of how token probabilities vary with reasoning correctness in R1-32B (\ie DeepSeek-R1-distill-Qwen-32B~\citep{deepseek}) and QwQ-32B~\citep{qwq}, both of which were likely to be post-trained over Qwen2.5-32B~\citep{qwen2.5}.
For both models, we report discourse tokens whose probabilities are significantly different when the answer is correct versus incorrect, as determined by a \textit{t}-test ($p<0.05$).
As shown, some tokens exhibit large differences (\eg ``wait'' in R1-32B, ``alternatively'' in QwQ-32B), while others show relatively smaller gaps (\eg ``so'' in R1-32B, ``let'' in QwQ-32B). 
Moreover, the same token (\eg ``the'') can behave differently across two models, suggesting variability in token-level signals. 
We define discourse tokens with $\Delta(t){>}0$ and $\Delta(t){<}0$ as correct- and incorrect-associated tokens, respectively.

Importantly, tokens with $\Delta(t)<0$, such as ``wait'' and ``alternatively'', commonly occur more frequently in incorrect trajectories, suggesting that they are employed to revise erroneous reasoning paths toward correct ones. This may reflect training data in which these discourse tokens appear at key transition points that redirect trajectories toward correct outcomes. However, how many such transition-inducing examples are required during training remains underexplored.

\begin{wraptable}{r}{0.43\textwidth}
\vspace{-.5em}
\centering
\small
\setlength\tabcolsep{8pt}
\begin{tabular}{r|ll}
\toprule
\textbf{Model} & \textbf{Pearson} & \textbf{Spearman} \\
\midrule
R1-32B & 0.9899$^{**}$ & 1.0000$^{***}$ \\
QwQ-32B & 0.8647$^{*}$ & 0.8286$^{*}$ \\
s1.1-32B & 0.9076$^{*}$ & 1.0000$^{***}$ \\
s1-32B & 0.7971 & 0.9000$^{*}$ \\
\bottomrule
\end{tabular}
\caption{
\textbf{Correlation between correct–incorrect token probability 
gap and model confidence}. Both Pearson and Spearman correlations are reported (*: $p{<}0.05$, **: $p{<}0.01$, ***: $p{<}0.001$).
}
\label{tab:correlation_difference}
\end{wraptable}

\noindent\textbf{Mean token probability correlates with confidence.} To further justify the effectiveness of our metric, 
we examine how token-level probability correlates to model confidence. We compute the self-reported confidence per trajectory following the design proposed by \citet{yoon2025reasoning} (see details in Appendix \ref{sec:app-1}). For each trajectory, the model confidence is compared with the trajectory-wise correct–incorrect token probability gap,
computed using \cref{eq:mtp} at the single-trajectory level as the difference between the summed probabilities of correct- and incorrect-associated tokens. This formulation allows for a more direct comparison between model confidence and token-level signals on each side.

As shown in \cref{tab:correlation_difference}, both Pearson and Spearman coefficients show strong positive correlations across all models, indicating that larger gaps
are aligned with higher confidence. 
These findings indicate that token-level probability patterns not only reflect correctness but are also significantly associated with the self-reported confidence of the model, supporting the reliability of our token-level analyses as meaningful behavioral indicators. We argue that the correlation between higher token probability and greater confidence suggests that the model can assess, at the token level, whether a trajectory is progressing correctly. This, in turn, indicates that discourse tokens act as token-level signals.


\subsection{Discourse tokens across setups and scales}

\noindent\textbf{Training recipes change discourse tokens.} 
We compare correct- and incorrect-associated tokens across models under the shared architectural baseline (Qwen2.5).
\cref{tab:models} presents the lists of correct-/incorrect-associated tokens across different models. 
First of all, tokens with negative or contrastive roles, such as ``however'', ``but'', ``consider'', and ``alternatively'', tend to appear as incorrect-associated tokens.  
In contrast, tokens that convey positive progression, such as ``therefore'', ``so'', and ``let'', are typically included among correct-associated tokens.  

The associated tokens vary across the models: R1-32B, QwQ-32B, and s1.1-32B, which all share the same base architecture, Qwen2.5-32B, yet their signals diverge depending on the training setups. s1.1-32B, trained on DeepSeek-R1 trajectories, shows strong similarity to R1-32B.
Similarly, QwQ-32B, though post-trained on the same backbone, shows further deviations due to its extensive combination of reinforcement learning and SFT. 

\begin{table*}[t]
\centering
\scriptsize
\begin{minipage}[t]{0.48\linewidth}
\centering
\begin{tabular}{l|c|l}
\toprule
\textbf{Model} & \textbf{Type} & \textbf{Associated Tokens} \\
\midrule
\multirow{2}{*}[-2pt]{R1-32B} & Correct   & { I, Therefore, The, Let, First} \\
\noalign{\vskip2pt}
\cline{2-3}
\noalign{\vskip2pt}
                        & Incorrect & Wait, Alternatively, But, In, If \\
\midrule
\multirow{2}{*}[-2pt]{QwQ-32B} & Correct   & I, Therefore, So, First, Now \\
\noalign{\vskip2pt}
\cline{2-3}
\noalign{\vskip2pt}
                         & Incorrect & This, If, In, The, Alternatively \\
\midrule
\multirow{2}{*}[-2pt]{s1.1-32B} & Correct   & Therefore, So, First, Now, Thus \\
\noalign{\vskip2pt}
\cline{2-3}
\noalign{\vskip2pt}
                          & Incorrect & Alternatively, Wait, Given, If, In \\
\bottomrule
\end{tabular}
\subcaption{Models post-trained on Qwen2.5-32B}
\label{tab:models}
\end{minipage}
\hfill
\begin{minipage}[t]{0.47\linewidth}
\centering
\begin{tabular}{c|c|l}
\toprule
\textbf{Model} & \textbf{Type} & \textbf{Associated Tokens} \\
\midrule
\multirow{2}{*}[-2pt]{32B} & Correct   & I, Therefore, The, Let, First \\
\noalign{\vskip2pt}
\cline{2-3}
\noalign{\vskip2pt}
                     & Incorrect & Wait, Alternatively, But, In, If \\
\midrule
\multirow{2}{*}[-2pt]{14B} & Correct   & Therefore, Let, Now, So, I \\
\noalign{\vskip2pt}
\cline{2-3}
\noalign{\vskip2pt}
                     & Incorrect & Wait, Alternatively, But, In, If \\
\midrule
\multirow{2}{*}[-2pt]{7B}  & Correct   & The, Now, Therefore, Let, I \\
\noalign{\vskip2pt}
\cline{2-3}
\noalign{\vskip2pt}
                     & Incorrect & Wait, Alternatively, Hmm, But, In \\
\bottomrule
\end{tabular}
\subcaption{Varying scales of R1}
\label{tab:scales}
\end{minipage}
\caption{\textbf{Correct-/incorrect-associated tokens across models and scales.} Three models at the same scale: \{R1-32B, QwQ-32B, s1.1-32B\} and three scales of R1-\{7B, 14B, 32B\} under a consistent training method are employed.
}
\label{tab:correct_tokens}
\end{table*}

\noindent\textbf{Model scales do not change discourse tokens.} We now control the fixed training strategy and compare three models of different sizes from the same series, R1-\{7B, 14B, 32B\}, as shown in \cref{tab:scales}. Interestingly, the sets of correct- and incorrect- associated tokens remain consistent across scales, with only minor variations in their relative importance ($|\Delta(t)|$). This consistency suggests that token-level patterns are largely independent of model capacity and are instead governed by the shared training recipe, such as the training corpus. Summing up the results in \cref{tab:correct_tokens}, they further suggest that the post-training corpus can trigger discourse tokens that revise reasoning trajectories, largely independent of model size. Notably, even a small-scale corpus such as s1.1 seemingly exhibits this effect to some extent.


\noindent\textbf{Summary of our findings.} 
We investigate token probabilities in models, focusing on which tokens serve as indicators of reasoning success and failure. 
Our analysis shows that progression tokens (\eg ``therefore'', ``so'') are consistently associated with correctness, while contrastive tokens (\eg ``wait'', ``alternatively'') are associated with incorrectness (Table \ref{tab:deepseek_tokens}). Since token-level probability patterns involved in decision-making correlate strongly with model confidence (\cref{tab:correlation_difference}), this supports the reliability of the proposed metric and provides insight into the behavior of individual tokens with respect to trajectory success or failure.
Additionally, the discourse tokens vary across training recipes but remain stable across model sizes (\cref{tab:correct_tokens}), suggesting that they are influenced more by training recipes than by capacity. 
    
Interestingly, in both R1-32B and s1.1-32B, which are trained on DeepSeek-R1 trajectories with the same backbone, ``wait'' consistently emerges as an incorrect-associated token. 
As observed in the probability of s1.1-32B in \cref{tab:s1_tokens}, R1-32B enjoys a larger $\Delta(t)$, but is noticeably weaker in s1.1-32B. This observation motivates us to see 1) a closer examination of why such closely related models differ in their use (for revising incorrect trajectories) of ``wait''; and 2) which signals are transferred through small-scale SFT concerning ``wait'', compared with larger datasets.



\section{``Wait'' in Reasoning Trajectories}
\label{sec:4}


The previous section highlighted that ``wait'' plays a key role in revising incorrect trajectories; we analyze it in depth and examine its correlation with reasoning success. We study how ``wait'' relates to reasoning progression and answer probability by truncating reasoning trajectories and prompting answers from incomplete reasoning.
We conduct analyses of R1-32B and s1.1-32B to examine how data-efficient SFT can approximate the effects of large-scale post-training. Our goal is not to establish causation, but rather to provide empirical observations that help explain why just post-training with a few yet curated samples can nonetheless exhibit similar trends.

\begin{figure*}[t]
\centering
\includegraphics[width=0.75\textwidth]{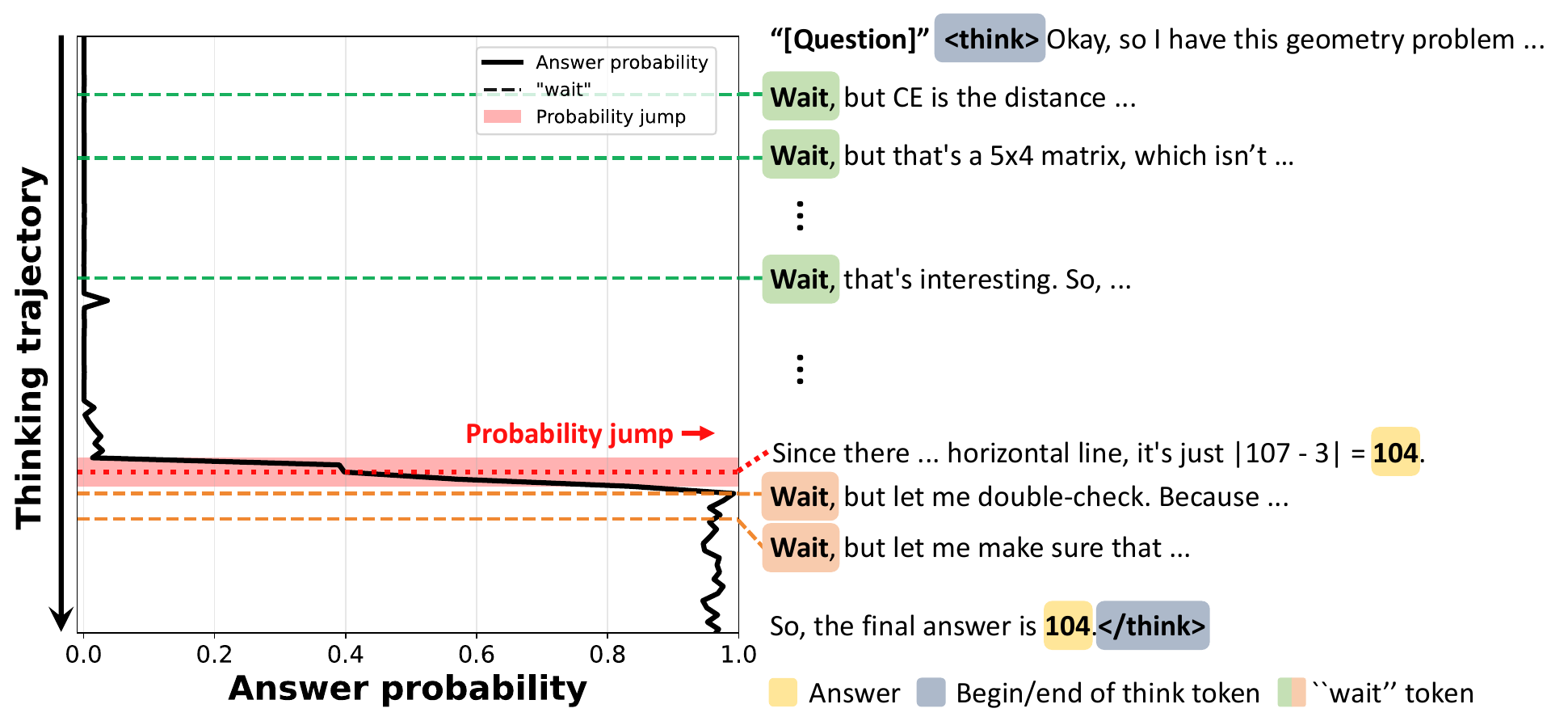}
\caption{\textbf{Answer probability} changes and \textbf{emergence of ``wait''} along the \textbf{reasoning trajectory}. Horizontal \textcolor{OliveGreen}{dashed }\textcolor{Peach} {lines} indicate where ``wait'' is generated, and the \textbf{Probability Jump} with the \textcolor{WildStrawberry}{red bar} marks the point of maximum increase. 
Expressions following ``wait'' differ depending on whether they occur before or after the probability jump: earlier {\setlength{\fboxsep}{0pt}\colorbox{YellowGreen}{Wait}} (\textit{rethink ``wait''}) more often extend the reasoning by attempting to revise the reasoning trace, whereas later {\setlength{\fboxsep}{0pt}\colorbox{Apricot}{Wait}} (\textit{recall ``wait''}) re-check even after a decision has already been made.
}
\label{fig:trajectory_ex}
\end{figure*}

\subsection{Probability jump and ``wait'' emergence}
We compute the (next-token's) answer probability as the probability that the model would generate the correct final answer. 
We compute this probability at intermediate points in the thinking trajectory by truncating it at fixed intervals of 10 tokens and prompting the model to directly generate the final answer. 
We insert the model-specific answer delimiter token, ``\texttt{\detokenize{<|im_start|>answer}}" for s1.1 and ``\texttt{\detokenize{</think>}}" for R1, followed by the prefix ``\texttt{\detokenize{Final answer: \boxed\{}}". This setup ensures that the model produces the final answer explicitly. 
We use AIME24~\citep{aime} for efficiency. 
Since all the answers in AIME24 are numerical, the probability of a correct answer is computed from the predicted distribution over digits. 
If the correct answer consists of multiple digits, we condition the generation by fixing the preceding digits and compute the probability of each subsequent digit sequentially. 
Formally, for an answer represented as a sequence of digits $a = (d_1, d_2, \ldots, d_T)$, the probability is given by $ P(a \mid \text{prompt}) \;=\; \prod_{t=1}^{T} P(d_t \mid d_{<t}, \text{prompt}).$

\noindent\textbf{How does the answer probability go?} 
The answer probability is expected to increase gradually as the reasoning trajectory unfolds. However, we observe that the probability often exhibits a sudden jump rather than a smooth trend (see Figure~\ref{fig:trajectory_ex} illustrating an example trajectory with a highlighted jump area). We call this phenomenon a \textit{probability jump} as the point in the trajectory where the increase in answer probability is maximized. 
We detect such a jump by sliding a window along each probability trajectory and computing, for each token position $t$, the difference between the average probability over the four preceding steps and the four following steps. 
The position $t$ that maximizes this difference is designated as the \textit{probability jump}. 
Notably, such sharp probability jumps are consistently observed whenever the model arrives at the correct answer.
Concurrently, \citet{tikhonov2026confidence} identify a similar phenomenon and exploit it for early stopping, whereas we use the jump as an analytical anchor for discourse-token dynamics.

\subsection{``wait'' is multifaceted}
Figure~\ref{fig:trajectory_ex} further illustrates that many ``wait'' tokens usually appear near a probability jump. 
To systematically analyze this trend, we classify every occurrence of ``wait” in the reasoning trajectory relative to the jump point. As discussed earlier, ``wait”  frequently appears in incorrect trajectories to revise them toward correct ones; moreover, it is also commonly observed in correct trajectories (\cref{tab:correct_tokens}). This trend suggests that ``wait”  plays a characteristic role within each trajectory, particularly with respect to answer probability - \ie ``wait'' is multifaceted.

\noindent\textbf{\Rethink and \recall.}
We decompose the role of ``wait'' into two categories: \rethink and \recall. 
A \rethink is defined as any ``wait'' token that appears before the probability jump, typically used to extend or reconsider the ongoing reasoning. 
Examples include phrases such as \textit{``Wait, but in this case ...''} or \textit{``Wait, but actually ...''}, where the token pushes the reasoning forward by exploring alternatives, and \textit{``Wait, that's interesting. So ...''}, where the token extends the reasoning by building on the current line of thought.
In contrast, a \recall is defined as any occurrence of ``wait'' after the jump, generally produced when the model has already reached the solution and is double-checking or summarizing its result. 
For instance, it appears in forms like \textit{``Wait, but let me double-check ...''} or \textit{``Wait, but let me think again ...''}, signaling verification or restatement. 
In other words, we label each ``wait'' token as either preceding the confidence leap (rethink) or following it (recall). We hereafter use the terms \textit{rethink} to denote the revision of incorrect trajectories, and \textit{recall} to denote the re-evaluation of corrected tokens after the transition.


\begin{figure*}[t]
\centering
\begin{subfigure}{0.35\textwidth}
    \includegraphics[width=\linewidth]{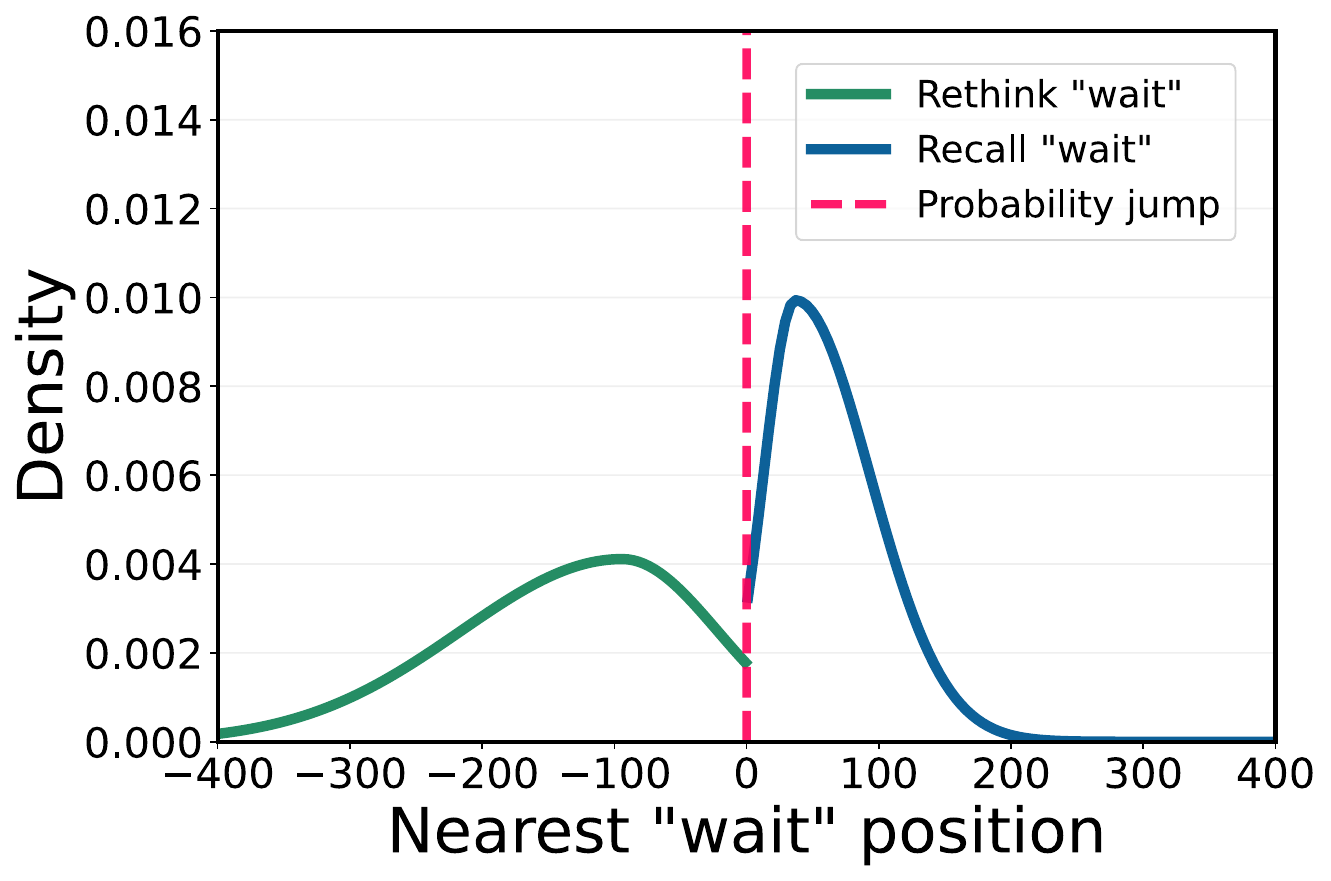}
    \caption{R1-32B}
    \label{fig:deepseek_spike}
\end{subfigure}
\quad \quad
\begin{subfigure}{0.35\textwidth}
    \includegraphics[width=\linewidth]{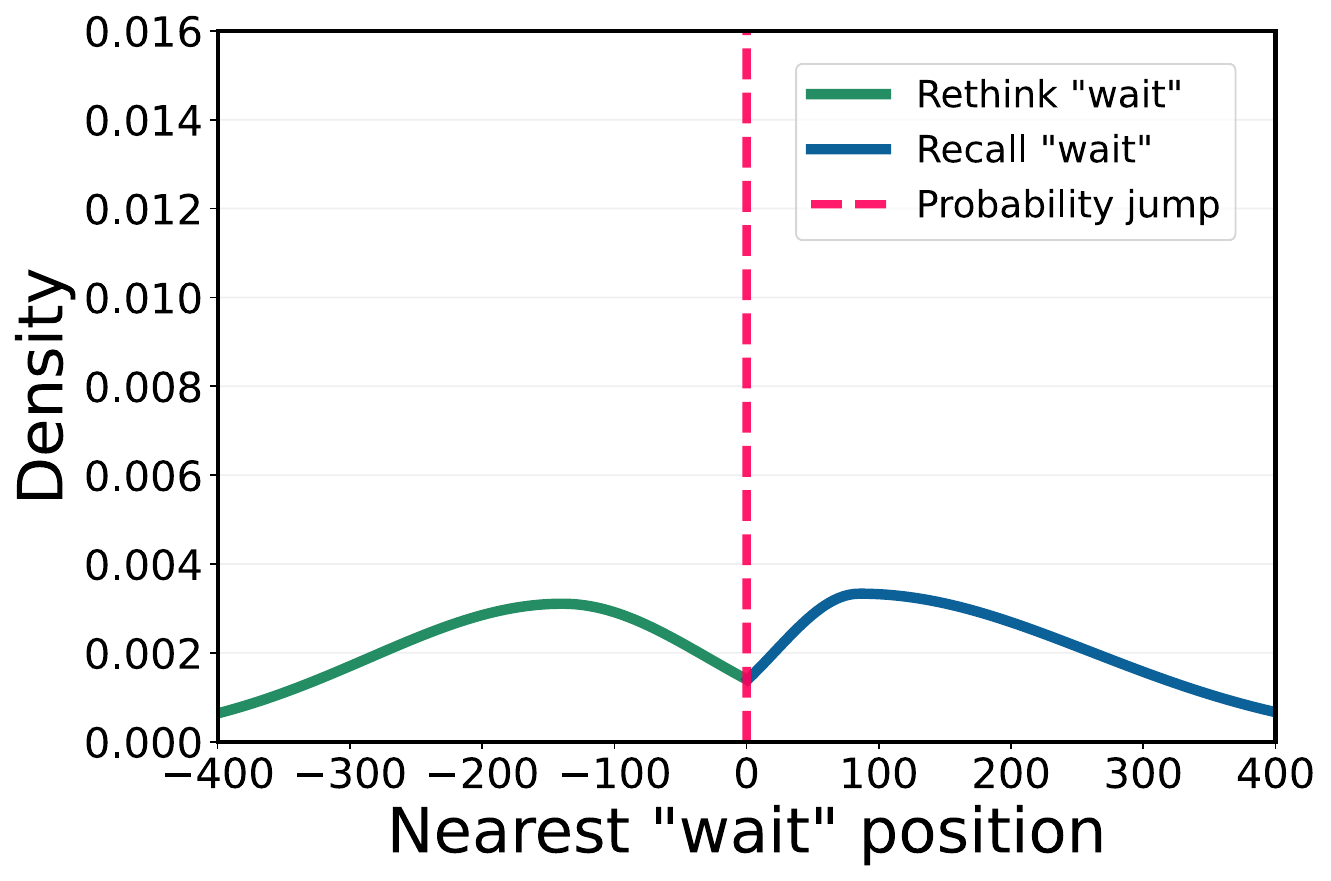}
    \caption{s1.1-32B}
    \label{fig:s11_spike}
\end{subfigure}
\caption{Distribution of the \textbf{relative positions of the nearest rethink and \recall token to the probability jump}. 
An asymmetric Gaussian curve is fitted to each distribution.}
\label{fig:wait_jump}
\end{figure*}

\subsection{Can Data-Efficient SFT Rival Large-Scale Post-Training?}
We analyze the patterns of \rethink and \recall tokens across two models: s1.1-32B and R1-32B to gain insights into data-efficient SFT relative to large-scale post-training.

\noindent\textbf{Distribution of \rethinks and \recalls and their functionality.}  
We compare \rethink and \recall around the probability jump for R1-32B and s1.1-32B. The distribution illustrated in \cref{fig:wait_jump} shows that, in the case of \rethinks (before the jump), they might trigger a reasoning step that makes the probability jump. After the jump, \recalls follow to review its reasoning and confirm the answer.
While the distributions of \rethink are similar across the two models, the \recalls show substantially different distributions between R1-32B and s1.1-32B.
This implies that although the s1.1-32B learns the position of \rethink from the small corpus enough to mimic R1-32B, yet the confirmation step via its revisiting process through \recall still differs, which is related to the gap in reasoning patterns and performance. 

Furthermore, for \rethinks, we argue that budget forcing~\citep{s1} works by intentionally inserting more ``wait" into the reasoning trace: when the number of wait tokens hits an internal threshold, a probability jump occurs. Although effective at test time, this may function as a temporary compensation by enriching many ``wait'' tokens. 


\begin{figure}[b]
\centering
\begin{subfigure}{0.35\columnwidth}
    \includegraphics[width=\linewidth]{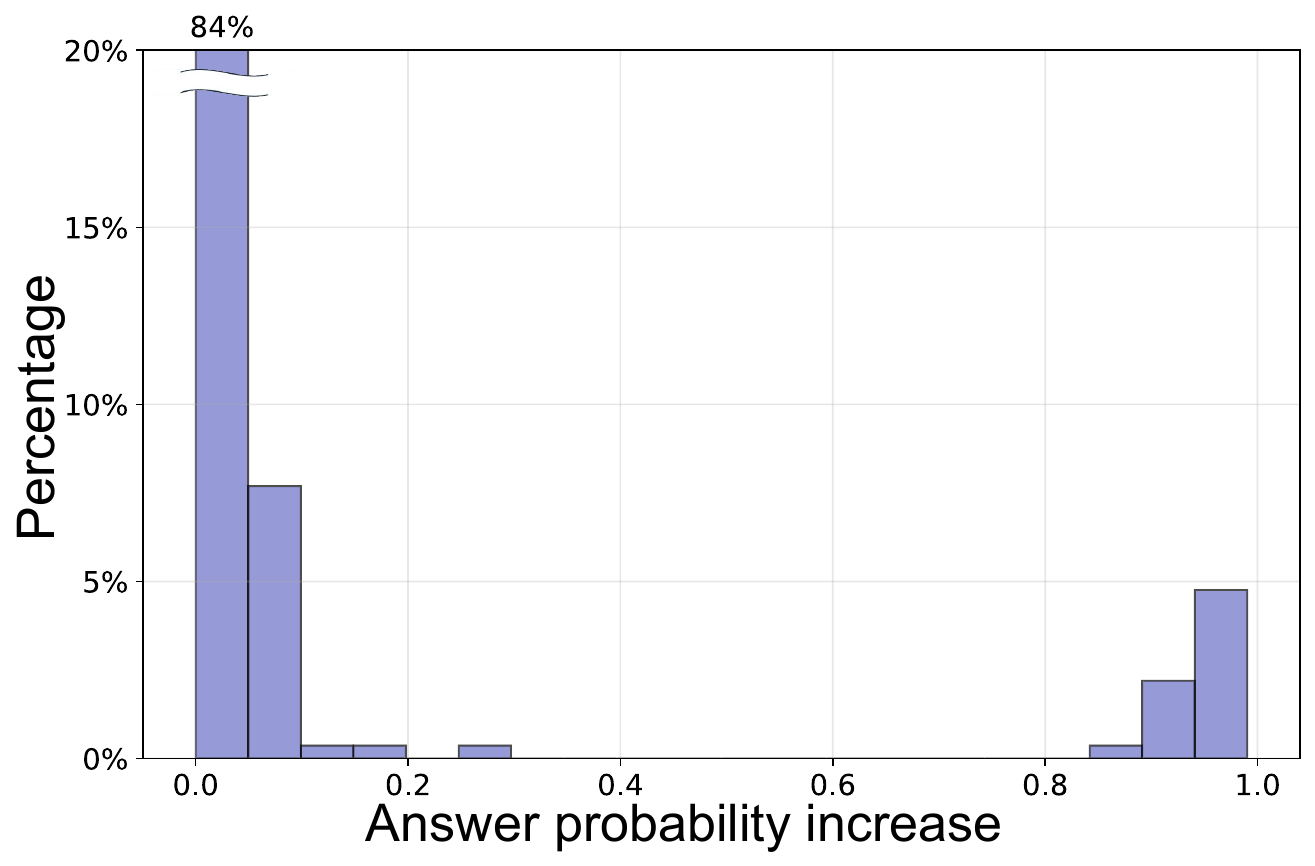}
    \caption{R1-32B}
    \label{fig:deepseek_prob_after_wait}
\end{subfigure}
\quad \quad
\begin{subfigure}{0.35\columnwidth}
    \includegraphics[width=\linewidth]{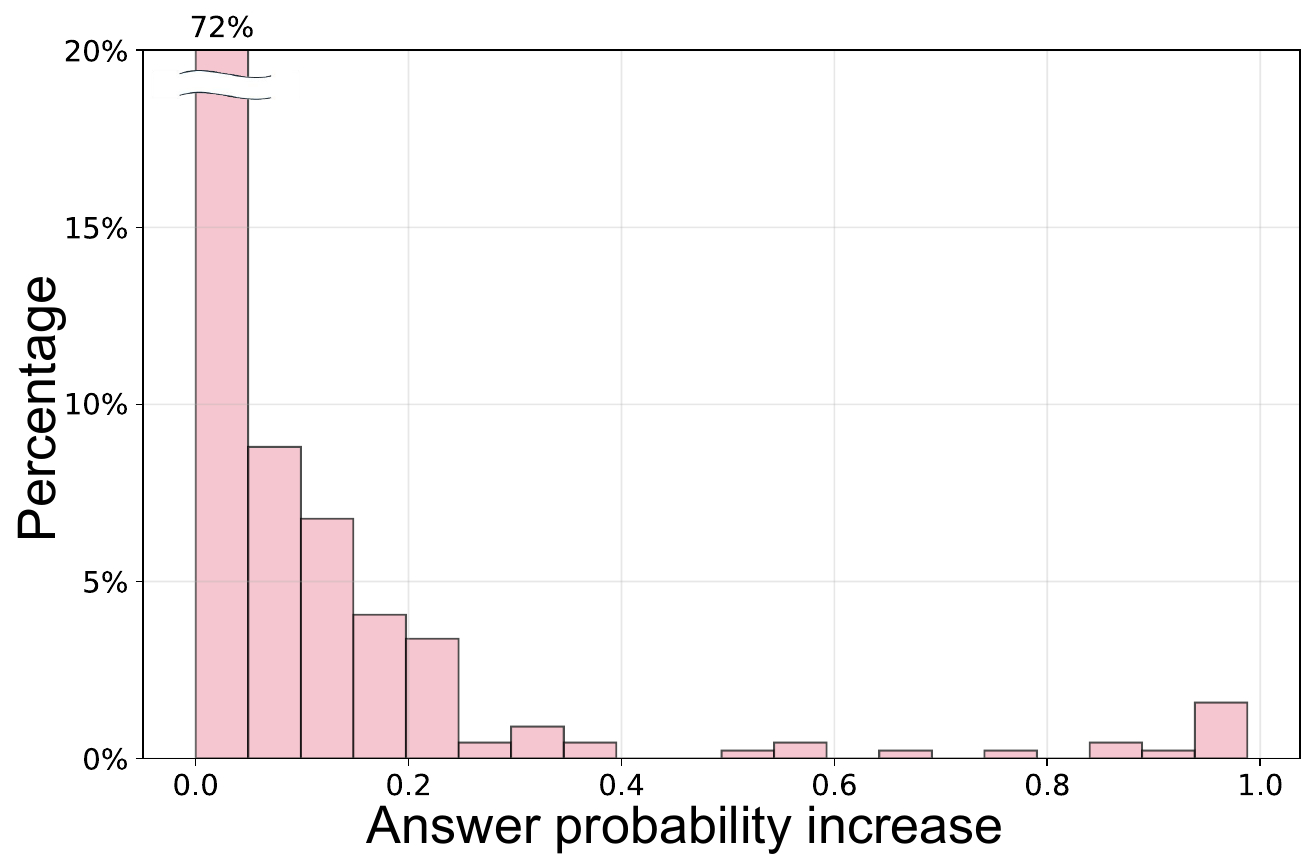}
    \caption{s1.1-32B}
    \label{fig:s11_prob_after_wait}
\end{subfigure}
\caption{\textbf{Distribution of answer probability increases} within a fixed window following the \rethink, shown for (a) R1-32B and (b) s1.1-32B.}
\label{fig:wait_increase}
\end{figure}

\begin{figure*}[t]
\centering
\begin{subfigure}{0.22\textwidth}
    \includegraphics[width=\linewidth]{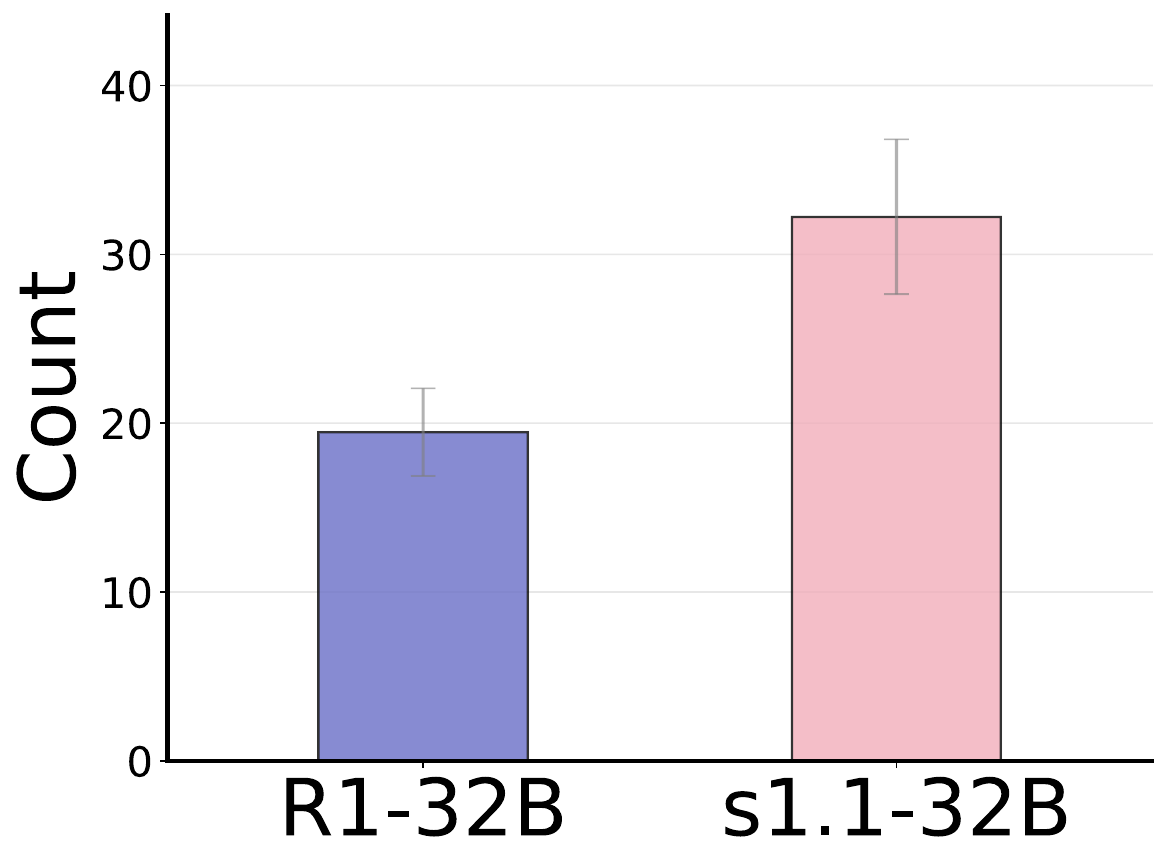}
    \caption{Total \rethink}
    \label{fig:stats_rethink}
\end{subfigure}
\hfill
\begin{subfigure}{0.22\textwidth}
    \includegraphics[width=\linewidth]{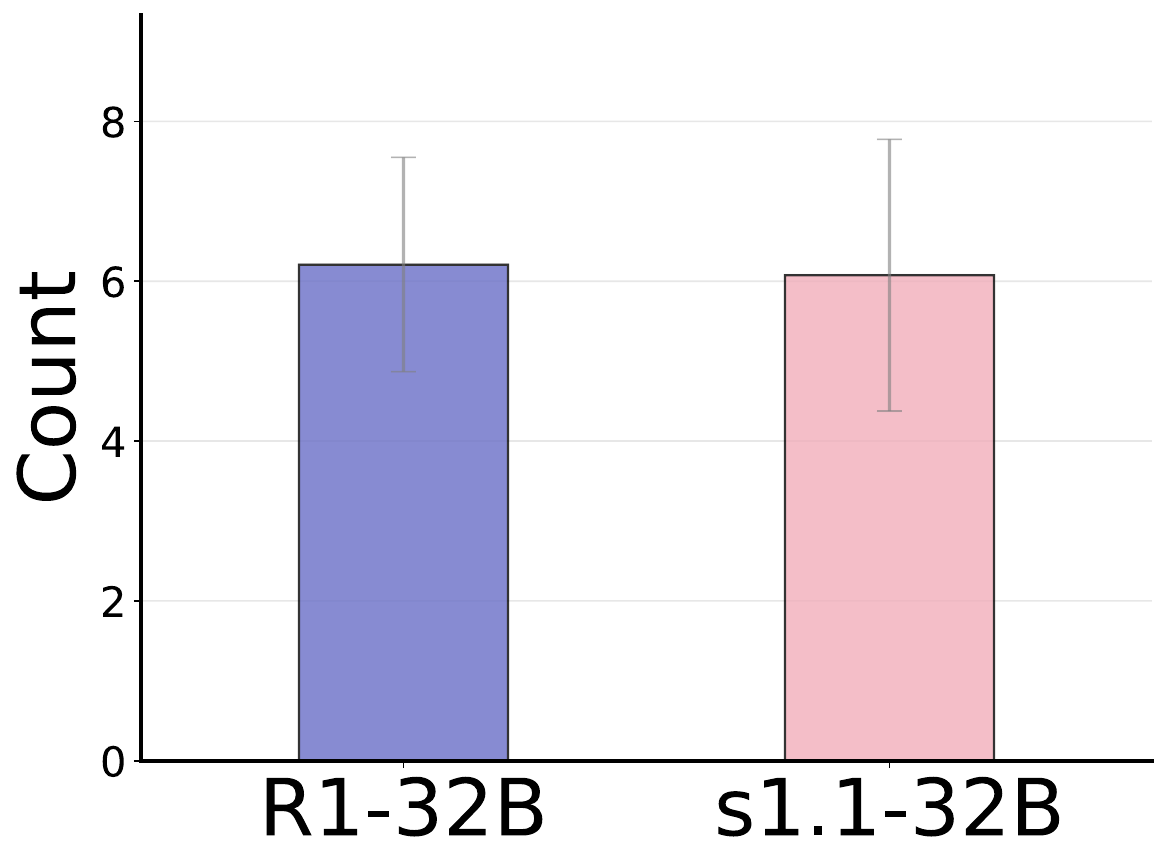}
    \caption{Total \recall}
    \label{fig:stats_recall}
\end{subfigure}
\hfill
\begin{subfigure}{0.22\textwidth}
    \includegraphics[width=\linewidth]{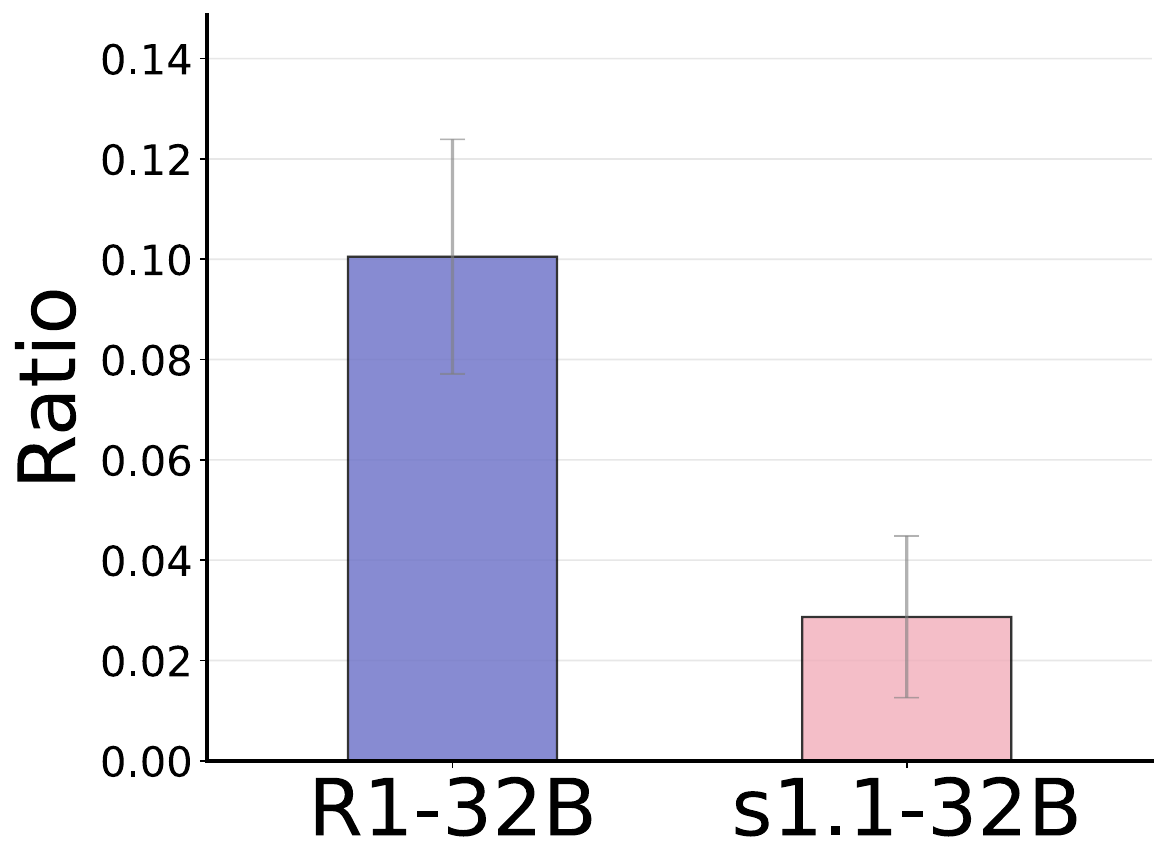}
    \caption{Success ratio}
    \label{fig:stats_success}
\end{subfigure}
\hfill
\begin{subfigure}{0.22\textwidth}
    \includegraphics[width=\linewidth]{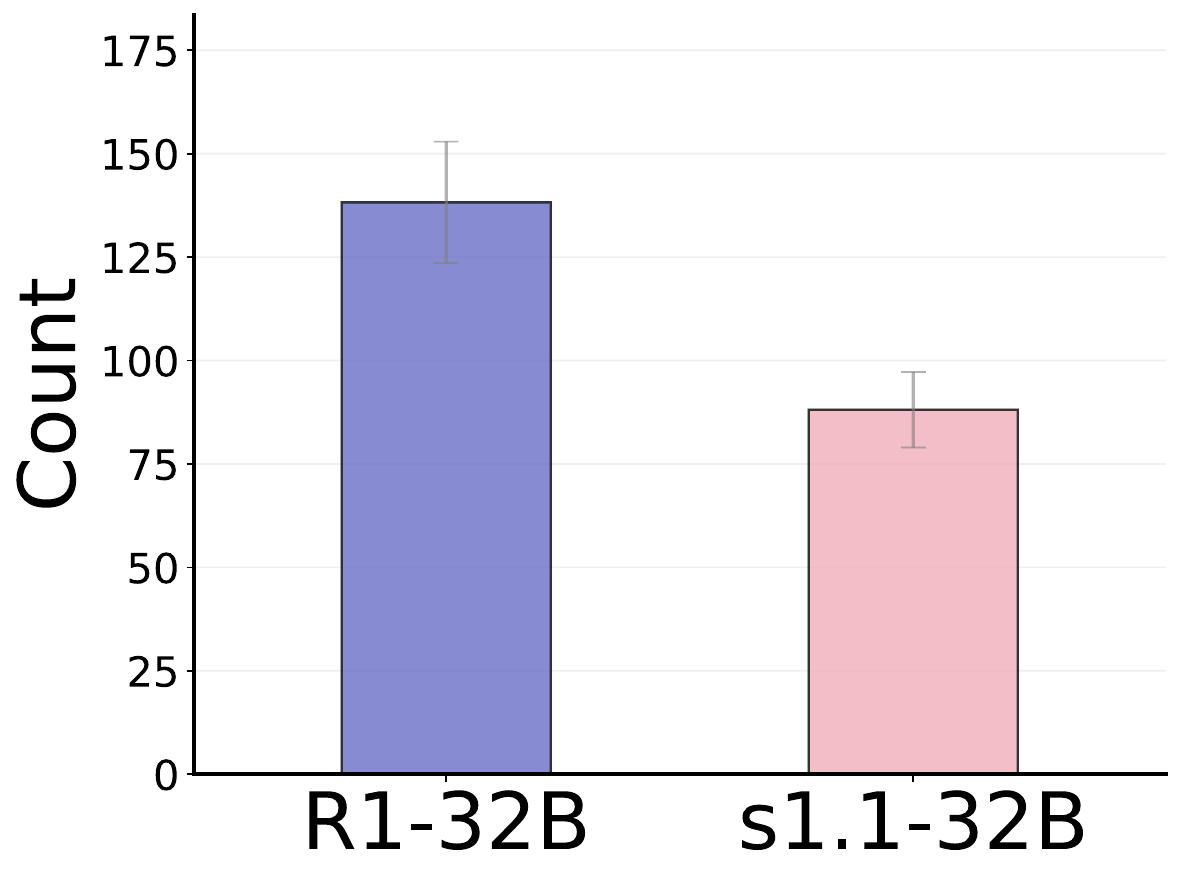}
    \caption{Incorrect samples}
    \label{fig:stats_incorrect}
\end{subfigure}
\caption{\textbf{Statistics of \textit{rethink} and \recall tokens} (more details beyond \cref{fig:wait_increase}). (a)–(d) show the number of \rethink and \recall tokens, the ratio of \rethink tokens followed by a significant probability increase, and the total number of ``wait'' tokens in incorrect trajectories, respectively.}
\label{fig:wait_stat}
\end{figure*}

\noindent\textbf{R1-32B: fewer  ``wait'' but reliable probability gains; s1.1-32B: many ``wait''  with low-probability exploration.} 
\cref{fig:wait_increase} shows the difference in answer probability increase distributions between R1-32B and s1.1-32B models. We report the maximum probability increases within a 384-token window after the ``wait'' token. 
For s1.1-32B, the answer probability increases are concentrated in the low-increase region (\ie $<$ 40\%) compared to R1-32B. A few \rethink make the probability jump close to 100\% (at the rightmost bin), but only in a small portion, 1.6\%.
Otherwise, R1-32B has a separate increase pattern: while it exhibits a reduced distribution in the low-increase region (\ie $<$ 40\%), it demonstrates a stronger increase at higher probabilities ($>$80\%).
\cref{fig:stats_success} reports the ratio of probability increase exceeds 80\% after applying question-level normalization. R1-32B \rethink is about four times more likely to make a probability jump compared to s1.1-32B.
Overall, these analyses indicate that ``wait'' tokens in R1-32B have superior effects compared with those in s1.1-32B.

More specifically, we quantify ``wait'' in R1-32B and s1.1-32B across all questions in \cref{fig:wait_stat}. As shown in \cref{fig:stats_rethink}, s1.1-32B uses approximately 1.5$\times$ more ``wait'' tokens than R1-32B in rethink cases, and these tokens are more widely distributed prior to the probability jump (see \cref{fig:wait_jump}). This suggests that s1.1 performs more trials to induce a probability jump, albeit with a low success ratio (\cref{fig:stats_success}) as also indicated in \cref{fig:wait_increase}. 
In contrast, the number of \recall is similar between the two models (\cref{fig:stats_recall}); however, their behavior near the probability jump differs as shown in \cref{fig:wait_jump}, suggesting that R1 performs rethinking more promptly. 

Finally, we visualize the number of ``wait'' in incorrect samples (\ie the reasoning trajectories where the final answer is incorrect) in \cref{fig:stats_incorrect}. 
Intriguingly, R1-32B generates more ``wait'' in the incorrect samples than s1.1, in contrast to its usage in correct samples. R1-32B appears to use more ``wait'' than s1.1-32B when it cannot find a path to the answer. This is likely due to larger training data, which provides more opportunities to learn where to place ``wait" tokens to revise incorrect trajectories into correct ones with higher confidence. While s1.1 exhibits similar effects, larger training corpora are evidently more beneficial for `effective' reasoning success and overall success ratio. Finally, we have focused on detailed analysis with ``wait'' due to its strong and consistent signal, similar (though weaker) trends would be observed for other discourse tokens such as ``alternatively''.



\noindent\textbf{Summary of our findings.}
We investigate the difference in the distribution of ``wait'' between s1.1-32B and R1-32B. R1-32B enjoys a more appropriate occurrence of ``wait'' compared to s1.1-32B in the following aspects. ``wait" emerges near the probability jump for R1-32B (\cref{fig:wait_jump}); 
efficient but high success ratio \rethink tokens (\cref{fig:stats_rethink} and \cref{fig:stats_success}); high ``wait'' counts for incorrect samples (\cref{fig:stats_incorrect}). 
Although the strength of data-efficient SFT lies in its ability to mimic large-scale post-training even with limited data, as evidenced by similar wait distributions and overall behavioral patterns, the results suggest that it offers limited advantages beyond reasoning performance. We believe that models trained on small-scale data rely on discourse-token-enriched training data, leading to superficial reasoning patterns driven by the high frequency of such tokens, rather than learning when and where they should arise.



\section{Exploratory Experiments}
\label{sec:5}
This section aims to reconfirm our claim and explore further applications.


\begin{table}[b]
\centering
\small
\begin{tabular}{l|c|ccc}
\toprule
& Baseline & Correct & Incorrect & All \\
\midrule
R1-32B     & 81.0\% & 80.2\% & 77.3\% & 78.5\% \\
QwQ-32B    & 74.7\% & 73.2\% & 70.1\% & 68.0\% \\
\midrule    
s1.1-32B   & 70.2\% & {74.9\%} & {71.2\%} &  {70.9\%} \\
\bottomrule
\end{tabular}
\vspace{.5em}
\caption{\textbf{Effect of associated-token suppression.} 
Baseline denotes no suppression; Correct, Incorrect, and All denote suppression of correct-/incorrect-/all associated tokens, respectively. Note that s1.1-32B shows divergent trends, where suppressing correct-associated tokens improves over baseline (74.9\% vs. 70.2\%). It suggests that data-efficient SFT enforces discourse-token patterns without acquiring the underlying robustness.
}
\label{tab:steering_results}
\end{table}

\noindent\textbf{Experiments on token suppression.}
Inspired by \cref{sec:3}, we perform an exploratory experiment to determine whether manipulating discourse tokens actually affects model performance.  
We examine this question by evaluating three models: R1-32B, QwQ-32B, and s1.1-32B across three settings: suppressing correct-associated tokens, incorrect-associated tokens (\cref{tab:deepseek_tokens}), and both groups simultaneously. We use AIME24, GPQA-Diamond, and MATH benchmarks and average the results. 


We observe a consistent pattern in large-scale post-trained models: while both types of tokens influence performance, tokens associated with incorrect trajectories play a more crucial role in maintaining reasoning ability. However, data-efficiently trained model (s1.1-32B) does not follow this trend. We conjecture that limited data forces the model to mimic these behaviors without acquiring sufficient robustness, so that removing the enforced signal can sometimes be beneficial, revealing a side effect.

\noindent\textbf{Introducing an ensembling method.}
Our analysis can also connect to select reliable samples across multiple trials to form an ensemble. 
Here, we instead utilize the correct–incorrect token probability gap, which exhibits a strong correlation with model confidence. 
We treat responses with low gap values as unreliable and exclude the bottom 20\% of samples based on this metric during ensembling.
All experiments are conducted on AIME24, where ensembling is performed over 32 sampled responses. We compare with: (1) majority voting, a general strategy for ensembling; (2) DeepConf~\citep{fu2025deep}, which recently introduced a group-level confidence for weighting or filtering answers from multiple trials.

\begin{table}[t]
\centering
\small
\begin{tabular}{l|cccc}
\toprule
 & Baseline & Maj. V. & DeepConf & Ours \\
\midrule
R1-32B & 50.0\% & \textbf{60.0\%} & \textbf{60.0\%} & \textbf{60.0\%} \\
QwQ-32B & 56.7\% & 63.3\% & 66.7\% & \textbf{70.0\%} \\
s1.1-32B & 43.3\% & 53.3\% & \textbf{56.7\%} & \textbf{56.7\%} \\
\bottomrule
\end{tabular}
\vspace{.5em}
\caption{\textbf{Ensemble using token-level signals.} Results are on the AIME24 benchmark. ``Maj. V.'' denotes the majority voting strategy; ours denotes an ensemble strategy based on the correct–incorrect token probability gap proposed in this work, respectively.}
\label{tab:ensemble_results}
\end{table}

\cref{tab:ensemble_results} presents the experimental results. All ensembling methods outperform the baseline (pass@1, without ensembling). Notably, our approach using the correct-incorrect token probability gap achieves the best performance across all three models. This observation suggests that our analysis can provide insights for enhancing model performance beyond indicating simple correlation with confidence or internal model signals.


\section{Conclusion}
\label{sec:1}

In this work, we have analyzed token-level signals to understand how large language models acquire and apply reasoning. Discourse tokens such as ``therefore'' and ``so'' are strongly linked to correct reasoning, whereas contrastive ones like ``wait'' and ``alternatively'' often accompany incorrect reasoning. 
These patterns remain stable across model sizes but vary with training recipes, suggesting that supervision plays a greater role than scale in shaping reasoning behavior. 
Further analysis of the ``wait'' token shows its dual role as both a trigger for probability shifts and a cue for self-checking. 
Our analysis also supports that data-efficient SFT captures token-level signals to approximate large-scale behavior, but only partially achieves the desired outcome. This, in turn, presumably explains why \citet{s1} benefits from inserting more discourse tokens like ``wait'' yet still exhibits performance gaps. 

\section*{Limitations}
While this study focuses on training strategies and model scales, our analyses primarily rely on open-source models from the Qwen series.
Future work could extend this framework to other base architectures such as LLaMA \cite{llama2} or Mixtral \cite{mixtral} to test its generality.
Because our method requires access to softmax outputs and full reasoning trajectories, it is less applicable to closed-source reasoning models such as GPT \cite{gpt4} or Gemini \cite{gemini}.
We also analyze three representative reasoning benchmarks that emphasize natural-language reasoning, where discourse markers naturally play a central role.
Code-generation settings \cite{quan2025codeelo,penedo2025codeforces,jain2024livecodebench} may exhibit different types of reasoning signals beyond discourse markers, which remain an interesting direction for future exploration.
Finally, our analyses are correlational; targeted interventions on individual "wait" tokens would be needed to causally verify their functional roles.

{\small
\bibliography{main}
\bibliographystyle{unsrtnat}
}

\appendix

\setcounter{table}{0}
\renewcommand{\thetable}{\Alph{table}}
\setcounter{figure}{0}
\renewcommand{\thefigure}{\Alph{figure}}
\setcounter{section}{0}
\renewcommand\thesection{\Alph{section}}

\newpage
\noindent{\Large \textbf{Appendix}} 
\label{sec:appendix}

\section{Experimental Settings}
\label{sec:app-1}

\noindent\textbf{Models.}
All models used in this study are publicly available open-source checkpoints released on HuggingFace under permissive licenses.
We use six reasoning-oriented LLMs: 
DeepSeek-R1-distill-Qwen-32B\footnote{\url{https://huggingface.co/deepseek-ai/DeepSeek-R1-Distill-Qwen-32B}}, 
DeepSeek-R1-distill-Qwen-14B\footnote{\url{https://huggingface.co/deepseek-ai/DeepSeek-R1-Distill-Qwen-14B}}, 
DeepSeek-R1-distill-Qwen-7B\footnote{\url{https://huggingface.co/deepseek-ai/DeepSeek-R1-Distill-Qwen-7B}}, 
s1.1-32B\footnote{\url{https://huggingface.co/simplescaling/s1.1-32B}}, 
s1-32B\footnote{\url{https://huggingface.co/simplescaling/s1-32B}}, 
and QwQ-32B\footnote{\url{https://huggingface.co/Qwen/QwQ-32B}}.
Licenses are MIT (DeepSeek-R1), Apache 2.0 (s1.1-32B and s1-32B), and Qwen Community License (QwQ-32B).
Each model is used solely for research purposes.

\noindent\textbf{Datasets and evaluation.}
We evaluate models on three reasoning-focused benchmarks: AIME24\footnote{\url{https://huggingface.co/datasets/simplescaling/aime24_nofigures}}, GPQA-Diamond\footnote{\url{https://huggingface.co/datasets/Idavidrein/gpqa}}, and MATH-500\footnote{\url{https://huggingface.co/datasets/simplescaling/openaimath}}. 
Each dataset is used solely for evaluation purposes for scientific research.
All evaluations are performed using the \texttt{lm-evaluation-harness} codebase~\citep{eval-harness}\footnote{\url{https://github.com/EleutherAI/lm-evaluation-harness}} and the \texttt{s1} codebase\footnote{\url{https://github.com/simplescaling/s1}}, both of which were adapted for our analysis.

All of these models and datasets are utilized for studying the reasoning of LLMs.
The models (e.g., Qwen2.5, DeepSeek-R1, and QwQ series) were pretrained on multilingual corpora that include both English and Chinese data.
However, all analyses and evaluations in this study were conducted exclusively on English-language benchmarks (AIME24, GPQA-Diamond, and MATH-500).

\noindent\textbf{Chat template for self-reported confidence.}
To obtain self-reported confidence in \cref{sec:3}, we use the chat template shown below, following the design proposed by \citet{yoon2025reasoning}.

\begin{chattemplatebox}
First, solve the following math problem efficiently and clearly.\\
Then, thoroughly assess your confidence in that answer by evaluating your thinking process so far.\\
Finally, classify your confidence into one of the following classes based on how likely your answer is to be correct:

- "Almost no chance" (0.0–0.1) \\
- "Highly unlikely" (0.1–0.2) \\
- "Chances are slight" (0.2–0.3)\\
- "Unlikely" (0.3–0.4)\\
- "Less than even" (0.4–0.5)\\
- "Better than even" (0.5–0.6)\\
- "Likely" (0.6–0.7)\\
- "Very good chance" (0.7–0.8)\\
- "Highly likely" (0.8–0.9)\\
- "Almost certain" (0.9–1.0)\\

Each category reflects the probability that your answer is correct.

The last line of your response should be of the following format: \detokenize{Therefore, the final answer is: $\boxed{{ANSWER}}$, Confidence: } \$CLASS. I hope it is correct.
(without quotes) where ANSWER is just the final number or expression that solves the problem and CLASS is one of the names (only the names without the probability ranges) of the classes above.
Think step by step before answering.\texttt{\textbackslash n\textbackslash n}

\end{chattemplatebox}

\section{Details on computing token probabilities}
\label{sec:app-2}

In \cref{sec:3}, we extract token-level signals through the average probability of tokens after ``\textbackslash n\textbackslash n''.
In practice, we store the top-20 logits at each step and restrict our analysis to tokens whose average generation probability exceeds 0.02 and that appear on average more than 20 times per question, ensuring statistical reliability.
We merge the probabilities of tokens that share the same semantic context but differ in surface form due to capitalization or leading whitespace (i.e., ``Wait'', ``wait'', `` Wait'', and `` wait'').
We conduct experiments on 30 questions from AIME24 \citep{aime}, 100 questions from GPQA-D \citep{gpqa}, and 100 questions from MATH-500 \citep{math}.  

We also consider calculating token probabilities at different token positions, not only after ``\textbackslash n\textbackslash n''. When token probabilities are computed after a dot or across all positions, the associated tokens for R1-32B in \cref{tab:deepseek_tokens} remain consistent. However, the strength of the signals is weakened. For example, the average true-associated probability $\bar p_{\text{true}}(t)$ of ``Wait'' is 15.4\% when considering positions after ``\textbackslash n\textbackslash n'', but it decreases to 2.15\% when considering positions after a dot and to 0.03\% when considering all positions. Furthermore, comparatively uninformative tokens, such as ``\$'' or ``='', appear to exhibit spurious signals due to problem-specific biases. Since our goal is to focus on discourse markers, we define token-level signals to be computed only at positions following ``\textbackslash n\textbackslash n''.

\begin{table}[h]
\centering
\small
\centering
\begin{tabular}{c r r r}
\toprule
\textbf{Token} & $\bar p_{true}(t)$ & $\bar p_{false}(t)$ 
& \multicolumn{1}{c}{$\Delta(t)$} \\
\midrule
Therefore      & 5.9\%  & 2.8\%  & +3.1\%  \\
So             & 5.9\%  & 3.2\%  & +2.7\%  \\
First          & 3.7\%  & 1.7\%  & +2.0\%  \\
Now            & 1.9\%  & 0.9\%  & +1.0\%  \\
Thus           & 1.4\%  & 0.7\%  & +0.7\%  \\
\midrule
In             & 0.6\%  & 1.1\%  & -0.6\%  \\
If             & 0.9\%  & 1.9\%  & -1.0\%  \\
Given          & 0.8\%  & 2.2\%  & -1.3\%  \\
Wait           & 8.2\%  & 10.5\% & -2.3\%  \\
Alternatively  & 10.3\% & 18.4\% & -8.1\%  \\
\bottomrule
\end{tabular}
\vspace{.5em}
\caption{\textbf{Discourse tokens with statistically significant differences}
(\textit{t}-test, $p<0.05$) 
between correct and incorrect trajectories for \textbf{S1.1-32B}. 
Tokens are sorted by $\Delta(t)$.}
\label{tab:s1_tokens}
\end{table}

\begin{table}[h]
\centering
\small
\setlength\tabcolsep{10pt}
\begin{tabular}{r|ll}
\toprule
\textbf{Model} & \textbf{Per-trace} & \textbf{Group} \\
\midrule
R1-32B & 0.5802$^{***}$ & 0.5993$^{***}$ \\
QwQ-32B & 0.7200$^{***}$ & 0.6585$^{***}$ \\
s1.1-32B & 0.6896$^{***}$ & 0.6092$^{***}$ \\
s1-32B &  0.5312$^{***}$ & 0.5019$^{**}$ \\
\bottomrule
\end{tabular}
\vspace{0.5em}
\caption{
\textbf{Correlation between the correct-incorrect token probability gap and model confidence based on log probability}. ``Per-trace'' denotes the confidence averaged over a single trajectory, while ``Group'' refers to the log probability-based confidence following DeepConf \citep{fu2025deep}.
Pearson correlations are reported (*: $p<0.05$, **: $p<0.01$, ***: $p<0.001$).}
\label{tab:correlation_log}
\end{table}

\section{Token-level signals and model confidence}
\label{sec:app-2-1}

In \cref{sec:3}, we obtain model confidence using self-reported confidence. Model confidence can also be estimated by the average log probability over a single generated trace (per-trace confidence), as used in~\citet{farquhar2024detecting}, or by group confidence, as suggested in~\citet{fu2025deep}. In this analysis, we follow DeepConf-low and compute group confidence using the bottom 10\% local token groups within a trajectory.
\cref{tab:correlation_log} presents the correlation results between our token-probability–based signals, derived from correct- and incorrect-associated tokens, and model confidence measured by log probability. Although the correlation is weaker than that observed with self-reported confidence, which exhibits clearer stratification due to its discrete confidence levels, the correlation remains statistically meaningful.

\section{Details on token suppression}
\label{sec:app-3}

For token suppression, we apply a masking strategy that prevents the generation of the correct- or incorrect-associated tokens listed in Table~\ref{tab:correct_tokens} during decoding, as a form of token-level steering.  
Each setting is run with a temperature of 0.6 for three trials, and we report the averaged results.

\section{The use of LLMs}
\label{sec:app-llm}
We use large language models to construct a small-scale SFT dataset by paraphrasing an existing dataset. We also use them for minor language refinement, such as improving fluency and clarity. They were not involved in any aspect of the study's design, analysis, or interpretation, and all research findings are entirely our own.

\end{document}